\documentclass[10pt,twocolumn,letterpaper]{article}

\usepackage{wacv}
\usepackage{times}
\usepackage{epsfig}
\usepackage{graphicx}
\usepackage{amsmath}
\usepackage{amssymb}
\usepackage{color}
\usepackage{booktabs}
\usepackage{authblk}
\usepackage[pagebackref=true,breaklinks=true,letterpaper=true,colorlinks,bookmarks=false,citecolor=blue,linkcolor=blue]{hyperref}
\wacvfinalcopy % *** Uncomment this line for the final submission

\def\wacvPaperID{108} % *** Enter the wacv Paper ID here

\ifwacvfinal\fi
\begin{document}

%%%%%%%%% TITLE
\title{Progressive Domain Adaptation for Object Detection}

% Authors at the same institution
% \author{First Author \hspace{2cm} Second Author \\
% Authors at different institutions
% \author{First Author \\
% \and
% Second Author \\

\makeatletter
\renewcommand\AB@affilsepx{\hspace{1mm} \protect\Affilfont}
\makeatother

\author[1]{Han-Kai Hsu}
\author[1]{Chun-Han Yao}
\author[2]{Yi-Hsuan Tsai}
\author[1]{Wei-Chih Hung}
\author[1]{\\Hung-Yu Tseng}
\author[3]{Maneesh Singh}
\author[1,4]{Ming-Hsuan Yang}
\affil[1]{UC Merced}
\affil[2]{NEC Labs America}
\affil[3]{Verisk Analytics}
\affil[4]{Google}

\maketitle
\ifwacvfinal\thispagestyle{empty}\fi
%%%%%%%%% ABSTRACT
\begin{abstract}
Recent deep learning methods for object detection rely on a large amount of bounding box annotations.
Collecting these annotations is laborious and costly, yet supervised models do not generalize well when testing on images from a different distribution.
Domain adaptation provides a solution by adapting existing labels to the target testing data.
However, a large gap between domains could make adaptation a challenging task, which leads to unstable training processes and sub-optimal results.
In this paper, we propose to bridge the domain gap with an intermediate domain and progressively solve easier adaptation subtasks.
This intermediate domain is constructed by translating the source images to mimic the ones in the target domain.
To tackle the domain-shift problem, we adopt adversarial learning to align distributions at the feature level.
In addition, a weighted task loss is applied to deal with unbalanced image quality in the intermediate domain.
Experimental results show that our method performs favorably against the state-of-the-art method in terms of the performance on the target domain.
\end{abstract}

%%%%%%%%% BODY TEXT
%figure: overall concept, 1. source to synthT (stlye) 2. synthT to target (content)
\vspace{-6mm}\section{Introduction}

Object detection is an important computer vision task aiming to localize and classify objects in images.
Recent advancement in neural networks has brought significant improvement to the performance of object detection~\cite{girshick2015fast, Ren_2017, Redmon_2016, Redmon_2017, Redmon2018YOLOv3AI, liu2016ssd}.
However, such deep models usually require a large-scale annotated dataset for supervised learning and do not generalize well when the training and testing domains are different.
For instance, domains can differ in scenes, weather, lighting conditions and camera settings.
Such domain discrepancy or domain-shift can cause unfavorable model generalization issues.
Although using additional training data from the target domain can improve the performance, collecting annotations is usually time-consuming and labor-intensive.

\begin{figure}[!htb]
		\begin{center}
        \includegraphics[width=\linewidth]{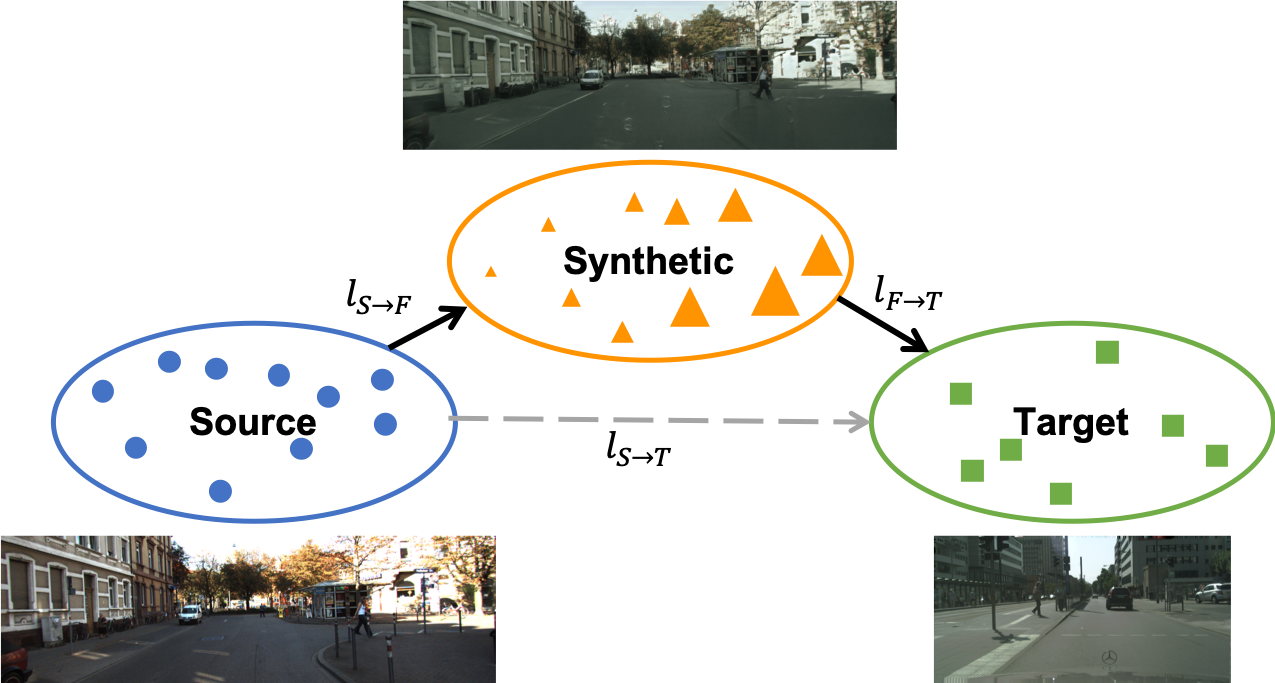}
		\end{center}
		\caption{
        An illustration of our progressive adaptation method.
        Conventional domain adaptation aims to solve domain-shift problem from source to target domain, which is denoted as $l_{\mathbb{S} \rightarrow \mathbb{T}}$.
        We propose to bridge this gap with an intermediate synthetic domain that allows us to gradually solve separate subtasks with smaller gaps (shown as $l_{\mathbb{S} \rightarrow \mathbb{F}}$ and $l_{\mathbb{F} \rightarrow \mathbb{T}}$).
        In addition, we treat each image in the synthetic domain unequally based on its quality with respect to the target domain, where the size of the yellow triangles stand for their weights (i.e., the closer to target, the higher of the weight).
		}
		\label{fig:overview}
\vspace{-2mm}
\end{figure}

Unsupervised domain adaptation methods address the domain-shift problem without using ground truth labels in the target domain.
Given the source domain annotations, the objective is to align source and target distributions in an unsupervised manner, so that the model can generalize to the target data without annotation effort.
Numerous methods are developed in the context of image classification \cite{Tzeng2014DeepDC, pmlr-v37-long15, Long2017DeepTL, sun2016deep, gopalan2011domain, Tzeng2017AdversarialDD, ganin2016domain, Bousmalis2016DomainSN}, while fewer efforts have been made on more complicated tasks such as semantic segmentation \cite{Hoffman2016FCNsIT, Tsai_adaptseg_2018} and object detection \cite{hoffman2014lsda, chen2018domain, Inoue_2018_CVPR}.
Such domain adaptation tasks are quite challenging as there usually exists a significant gap between source and target domains.

In this paper, we aim to ease the effort of aligning different domains.
Inspired by \cite{gopalan2011domain} which addresses the domain-shift problem via aligning intermediate feature representations, we utilize an intermediate domain that lies between source and target, and hence avoid direct mapping across two distributions with a significant gap.
Specifically, the source images are first transformed by an image-to-image translation network~\cite{CycleGAN2017} to have similar appearance as the target ones. We refer to the domain containing synthetic target images as the intermediate domain.
We then construct an intermediate feature space by aligning the source and intermediate distributions, which is an easier task than aligning to the final targets.
Once this intermediate domain is aligned, we use it as a bridge to further connect to the target domain.
As a result, via the proposed progressive adaptation through the intermediate domain, the original alignment between source and target domains is decomposed into two subtasks that both solve an easier problem with a smaller domain gap.

During the alignment process, since the intermediate space is constructed in an unsupervised manner, one potential issue is that each synthetic target image may contribute unequally based on the quality of the translation.
To reduce the outlier impact of the low-quality translated images, we propose a weighted version in our adaptation method, where the weight is determined based on the distance to the target distribution.
That is, an image closer to the target domain should be considered a more important sample.
In practice, we obtain the distance from the discriminator in the image translation model and incorporate it into the detection framework as a weight in the task loss.

We evaluate our method on various adaptation scenarios using numerous datasets, including KITTI \cite{Geiger2012CVPR}, Cityscapes \cite{Cordts2016Cityscapes}, Foggy Cityscapes \cite{SDV18} and BDD100k \cite{Yu2018BDD100KAD}.
We conduct experiments on multiple real-world domain discrepancy cases, such as weather changes, camera differences and the adaptation to a large-scale dataset.
With the proposed progressive adaptation, we show that our method performs favorably against the state-of-the-art algorithm in terms of accuracy in the target domain.
The main contributions of the work are summarized as follows:
1) we introduce an intermediate domain in the proposed adaptation framework to achieve progressive feature alignment for object detection,
2) we develop a weighted task loss during domain alignment based on the importance of the samples in the intermediate domain, and
3) we conduct extensive adaptation experiments under various object detection scenarios and achieve state-of-the-art performance.

%------------------------------------------------------------------------
\section{Related Work}
\paragraph{Object Detection.}
Recently, state-of-the-art object detection methods are predominately based on the deep convolutional neural networks~(CNNs).
These methods can be categorized into region proposal-based and single-shot detectors, depending on the network forwarding pipelines.
Region proposal-based methods~\cite{girshick2015fast, Ren_2017} perform prediction on a variable set of candidate regions.
Fast R-CNN~\cite{girshick2015fast} applies selective search~\cite{uijlings2013selective} to obtain region proposals, while Faster R-CNN~\cite{Ren_2017} proposes to learn a Region Proposal Network~(RPN) to accelerate the proposal generation process.
To further reduce the computational need of proposal generation, single-shot approaches~\cite{Redmon_2016, Redmon_2017, Redmon2018YOLOv3AI, liu2016ssd} employ a fixed set of predefined anchor boxes as proposals and directly predict the category and offsets for each anchor box.
Although these methods achieve state-of-the-art performance, such success hinges on the substantial amount of labeled training data which requires a high labor cost.
Also, these methods can overfit on the training domain, which makes them difficult to generalize to many real-world scenarios.
As a result, the vision community has recently started showing a great interest in employing domain adaptation techniques to object detection.

\begin{figure*}
		\begin{center}
        \includegraphics[width=\textwidth]{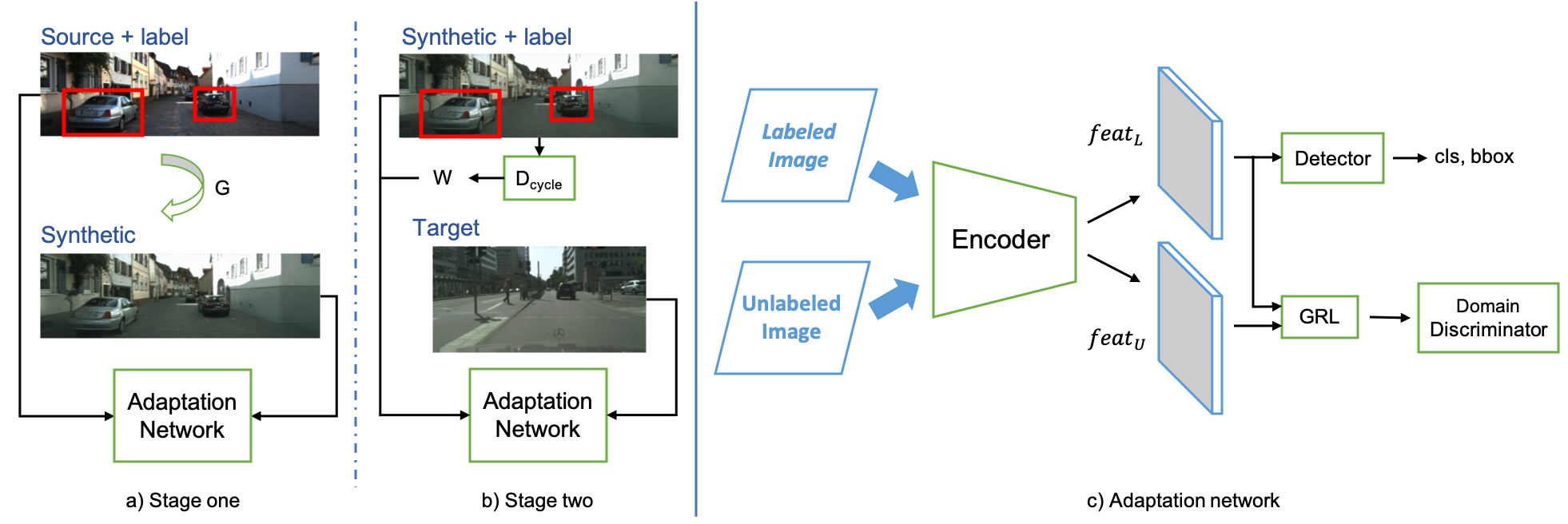}
		\end{center}
		\caption{
        The proposed progressive adaptation framework.
        The algorithm includes two stages of adaptation as shown in a) and b).
        In a), we first transform source images to generate synthetic ones by using the generator $G$ learned via CycleGAN \cite{CycleGAN2017}.
        Afterward, we use the labeled source domain and perform first stage adaptation to the synthetic domain.
        Then in b), our model applies a second stage adaptation which takes the synthetic domain with labels inherited from the source and aligns the synthetic domain features with the target distribution.
        In addition, a weight $w$ is obtained from the discriminator $D_{cycle}$ in CycleGAN to balance the synthetic image qualities in the detection loss.
        The overall structure of our adaptation network is shown in c).
        Labeled and unlabeled images are both passed through the encoder network $E$ to extract CNN features $feat_{L}$ and $feat_{U}$. 
        We then use them to: 1) learn supervised object detection with the detector network from $feat_{L}$, and 2) forward both features to GRL and a domain discriminator, learning domain-invariant features in an adversarial manner.
		}
		\label{fig:da_net}
\end{figure*}

\vspace{-4mm}\paragraph{Domain Adaptation.}
Domain adaptation techniques aim to tackle domain-shift between the source and target domains with unlabeled or weakly labeled images in the target domain.
In recent years, adversarial learning has played a critical role in domain adaptation methods.
Since the emergence of the Domain Adversarial Neural Network~(DANN)~\cite{ganin2016domain}, numerous works~\cite{Bousmalis2016DomainSN,Tzeng2017AdversarialDD,chen2018domain} have been proposed to utilize adversarial learning for the feature distribution alignment between two domains.
Furthermore, several methods attempt to perform alignment in the pixel space, based on the unpaired image-to-image translation approaches~\cite{CycleGAN2017}.
For image classification, PixelDA~\cite{bousmalis2017unsupervised} synthesizes additional images in the target domain by learning one-to-many mapping. 
For semantic segmentation, CyCADA~\cite{hoffman2017cycada} and AugGAN~\cite{huang2018auggan} both design a CycleGAN~\cite{CycleGAN2017}-like network to transform images from the source domain to the target one.
The transformed images are then treated as simulated training images for the target domain with the same label mapped from the source domain.
Instead of performing alignment in the feature/pixel space, Tsai~\etal~\cite{Tsai_adaptseg_2018, Tsai_DA4Seg_ICCV19} adopt adversarial learning in the structured output space for solving domain adaptation on semantic segmentation.

To address domain adaptation for object detection in a weakly-supervised manner, LSDA \cite{hoffman2014lsda} finetunes a fully-supervised classification model for object detection with limited bounding box resources. 
Alternatively, Naoto \etal~\cite {Inoue_2018_CVPR} train the network with synthetic data and finetune it with pseudo-labels in the target domain.
In an unsupervised domain adaptation setting, Chen~\etal~\cite{chen2018domain} propose to close the domain gap on both image level and instance level via adversarial learning.
%
%Zhu~\etal~\cite{Zhu_2019_CVPR} emphasize on aligning the discriminative regions, and hence propose to align the regions of interest using generative adversarial networks.
%
%Saito~\etal~\cite{Saito_2019_CVPR} propose novel approach which strongly align the low-level features and weakly align the high-level features by focusing on the globally similar images.
% \textcolor{red}{
To emphasize on matching local features, Zhu~\etal~\cite{Zhu_2019_CVPR} mines discriminative regions for alignment,
while Saito~\etal~\cite{Saito_2019_CVPR} focus on aligning local receptive fields at low-level features along with weak alignment on global regions.
On the other hand, Kim~\etal~\cite{kim2019diversify} utilize image translation network to generate multiple domains and use a multi-domain discriminator to adapt all domains simultaneously, but this method does not consider the distance between the generated ones and the final target.
% }

In this work, we observe that simply applying image translation without knowing the distance between each generated sample and the target domain may result in less effective adaptation.
%
% adversarial learning may not be sufficient since the distance between the source and target feature distributions is often too large for a one-step alignment.
%
To handle this issue, we first introduce an intermediate domain to reduce the effort of mapping two significantly different distributions and then adopt a two-stage 
alignment strategy with sample weights to account for the sample quality.

%-------------------------------------------------------------------------

\section{Progressive Domain Adaptation}
We propose to decompose the domain adaptation problem into two smaller subtasks, bridged by a synthetic domain sitting in between the source and target distribution.
Taking advantage of this synthetic domain, we adopt a progressive adaptation strategy which closes the gap gradually through the intermediate domain.
We denote the source, synthetic, and target domains as $\mathbb{S}$,~$\mathbb{F}$ and~$\mathbb{T}$, respectively.
The conventional adaptation from a labeled domain $\mathbb{S}$ to the unlabeled domain $\mathbb{T}$ is denoted as $\mathbb{S} \rightarrow \mathbb{T}$,
while the proposed adaptation subtasks are expressed as $\mathbb{S} \rightarrow \mathbb{F}$ and $\mathbb{F} \rightarrow \mathbb{T}$.
An overview of our progressive adaptation framework is shown in Figure \ref{fig:da_net}.
We discuss the details of the proposed adaptation network and progressive learning in the following sections.

\subsection{Adaptation in the Feature Space}
In order to align distributions in the feature space, we propose a deep model which consists of two components; a detection network and a discriminator network for feature alignment via adversarial learning.
\vspace{-2mm}\paragraph{Detection Network.} We adopt the Faster R-CNN \cite{Ren_2017} framework for the object detection task, where the detector has a base encoder network $E$ to extract image features.
Given an image $\mathbf{I}$, the feature map $E(\mathbf{I})$ is extracted and then fed into two branches: Region Proposal Network (RPN) and Region of Interest (ROI) classifier. We refer to these branches as the detector, which is shown in Figure~\ref{fig:da_net}.
To train the detection network, the loss function $\mathcal{L}_{det}$ is defined as:
\begin{equation} \label{eq:det_loss}
\mathcal{L}_{det}(E(\mathbf{I})) = \mathcal{L}_{rpn} + \mathcal{L}_{cls} + \mathcal{L}_{reg},
\end{equation}
where $\mathcal{L}_{rpn}$, $\mathcal{L}_{cls}$, and $\mathcal{L}_{reg}$ are the loss functions for the RPN, classifier and bounding box regression, respectively.
We omit the details of the RPN and ROI classifier here as we focus on solving the domain-shift We omit the details of the RPN and ROI classifier here as we focus on solving the domain-shift problem. The readers are encouraged to refer to the original paper \cite{Ren_2017} for further details.

\vspace{-2mm}\paragraph{Domain Discriminator.}
To align the distributions across two domains, we append a domain discriminator $D$ after the encoder $E$.
The main objective of this branch is to discriminate whether the feature $E(\mathbf{I})$ is from the source or the target domain.
Through this discriminator, the probability of each pixel belonging to the target domain is obtained as $\mathbf{P} = D(E(\mathbf{I})) \in \mathbb{R}^{H \times W}$.
We then apply a binary cross-entropy loss to $\mathbf{P}$ based on the domain label $d$ of the input image,
where images from the source distributions are given the label $d = 0$ and the target images receive label $d = 1$.
The discriminator loss function $\mathcal{L}_{disc}$ can be formulated as:
\begin{align}
\mathcal{L}_{disc}(E(\mathbf{I})) = - \sum_{h,w} \ \ &d \log \mathbf{P}^{(h,w)} \nonumber \\
&+ (1-d)  \log (1 - \mathbf{P}^{(h,w)}).
 \label{eq:obj}
\end{align}

\vspace{-2mm}\paragraph{Adversarial Learning.} 
Adversarial learning is achieved using the Gradient Reverse Layer (GRL) proposed in \cite{ganin2015unsupervised} to learn the domain-invariant feature $E(\mathbf{I})$.
GRL is placed in between the discriminator and the detection network, only affecting the gradient computation in the backward pass.
During backpropagation, GRL negates the gradients that flow through.
As a result, the encoder $E$ receives gradients that force it to update in an opposite direction which maximizes the discriminator loss.
This allows $E$ to produce features that fools the discriminator $D$ while $D$ tries to distinguish the domain of the features.
For the adaptation task $\mathbb{S} \rightarrow \mathbb{T}$, given source images $\mathbf{I}_\mathbb{S}$ and target images $\mathbf{I}_\mathbb{T}$, the overall min-max loss function of the adaptive detection model is defined as the following:
\begin{align}
  \min_{E} \max_{D} \mathcal{L}(\mathbf{I}_\mathbb{S}, \mathbf{I}_\mathbb{T}) = \mathcal{L}_{det}(\mathbf{I}_\mathbb{S}) + &\lambda_{disc} \big[ \mathcal{L}_{disc}(E(\mathbf{I}_\mathbb{S})) \nonumber \\
 & + \mathcal{L}_{disc}(E(\mathbf{I}_\mathbb{T})) \big],
  \label{eq:objective}
\end{align}
where $\lambda_{disc}$ is a weight applied to the discriminator loss that balances the loss. 

\subsection{Progressive Adaptation}
Aligning feature distributions between two distant domains is challenging, and hence we introduce an intermediate feature space to make the adaptation task easier.
That is, instead of directly solving the gap between the source and the target domains, we progressively perform adaptation to the target domain bridged by the intermediate domain.

\vspace{-2mm}\paragraph{Intermediate Domain.}
%We introduce an intermediate domain to assist our progressive adaptation.
%
The intermediate domain is constructed from the source domain images to synthesize the target distributions on the pixel-level.
We apply an image-to-image translation network, CycleGAN~\cite{CycleGAN2017} to learn a function that maps the source domain images to the target ones, and vice versa.
Since ground truth labels are only available in the source domain, we only consider the translation from source images to the target domain (i.e., synthetic target images) after training CycleGAN.

\begin{figure}
		\begin{center}
        \includegraphics[width=0.7\linewidth]{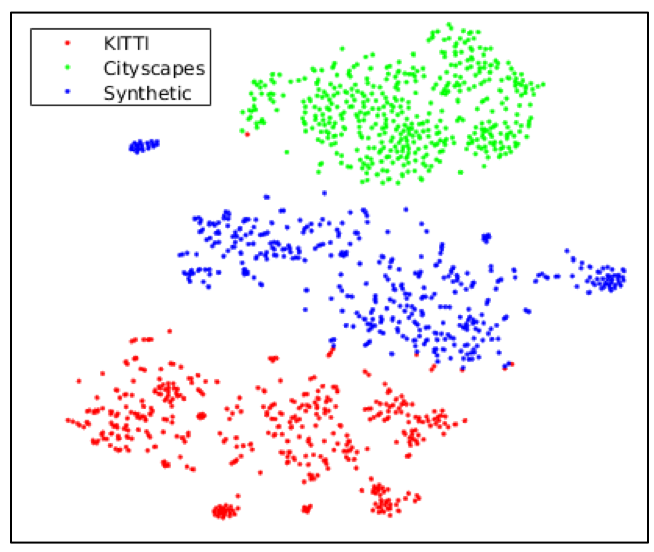}
		\end{center}
		\caption{
        Visualization of the feature distributions via t-SNE \cite{vanDerMaaten2008}, showing that our synthetic images serve as an intermediate feature space between the source and target distributions.
        Each dot represents one image feature extracted from $E$.
        We take 500 images from the Cityscapes validation set and 500 from the KITTI training set for comparison.
  		}
		\label{fig:tsne_before}
\vspace{-3mm}
\end{figure}

Synthetic target images have been utilized to assist with domain adaptation tasks \cite{bousmalis2017unsupervised, huang2018auggan, Inoue_2018_CVPR} as additionally augmented target training data.
Different from these approaches, we define this set of synthetic images as an individual domain $\mathbb{F}$ to connect the labeled domain $\mathbb{S}$ with the unlabled domain $\mathbb{T}$ via adversarial learning.
One motivation behind this is that the similarity between source domain $\mathbb{S}$ and $\mathbb{F}$ is the image content, only diverging in the visual appearances, while $\mathbb{F}$ and the target domain $\mathbb{T}$ are different in image details but have similar distributions on the pixel-level.
Consequently, this synthetic domain ``sits'' in between the source and target domains and thus can help reduce the adaptation difficulty of a large domain gap between $\mathbb{S}$ and $\mathbb{T}$.
Figure \ref{fig:tsne_before} is one example of feature space visualization using the KITTI and Cityscapes datasets.
This figure shows a distribution plot by mapping the features from $E(\mathbf{I})$ to a low dimensional 2-D space via t-SNE \cite{vanDerMaaten2008}.
The plot demonstrates that in the feature space, the synthetic domain $\mathbb{F}$ (blue) is located in between the KITTI (red) and Cityscapes (green) distributions.

\vspace{-2mm}\paragraph{Adaptation Process.}
Our domain adaptation network involves obtaining knowledge from a labeled source domain $\mathbb{S}$ then map that knowledge to an unlabeled target domain $\mathbb{T}$ by aligning the two distributions, solving the adaptation task $\mathbb{S} \rightarrow \mathbb{T}$, i.e., via \eqref{eq:objective} in this paper.
To take advantage of the intermediate feature space during alignment, our algorithm decomposes the problem into two stages: $\mathbb{S} \rightarrow \mathbb{F}$ and $\mathbb{F} \rightarrow \mathbb{T}$, as shown in Figure \ref{fig:da_net} a) and b).
At the first stage, we use $\mathbb{S}$ as the labeled domain, adapting to $\mathbb{F}$ without labels.
Due to the underlying similarity between $\mathbb{S}$ and $\mathbb{F}$ in image contents, the network focuses on aligning the feature distributions with respect to the appearance difference on the pixel-level.
After aligning pixel discrepancies between $\mathbb{S}$ and $\mathbb{F}$, we take $\mathbb{F}$ as the source domain for supervision and adapts to $\mathbb{T}$ as stage two in the proposed method.
During this step, the model can take advantage of the appearance-invariant features from the first step and focus on adapting the object and context distributions.
In summary, the proposed progressive learning separates the adaptation task into two subtasks and pays more attention to individual discrepancies during each adaptation stage.

\begin{figure}
		\begin{center}
        \includegraphics[width=\linewidth]{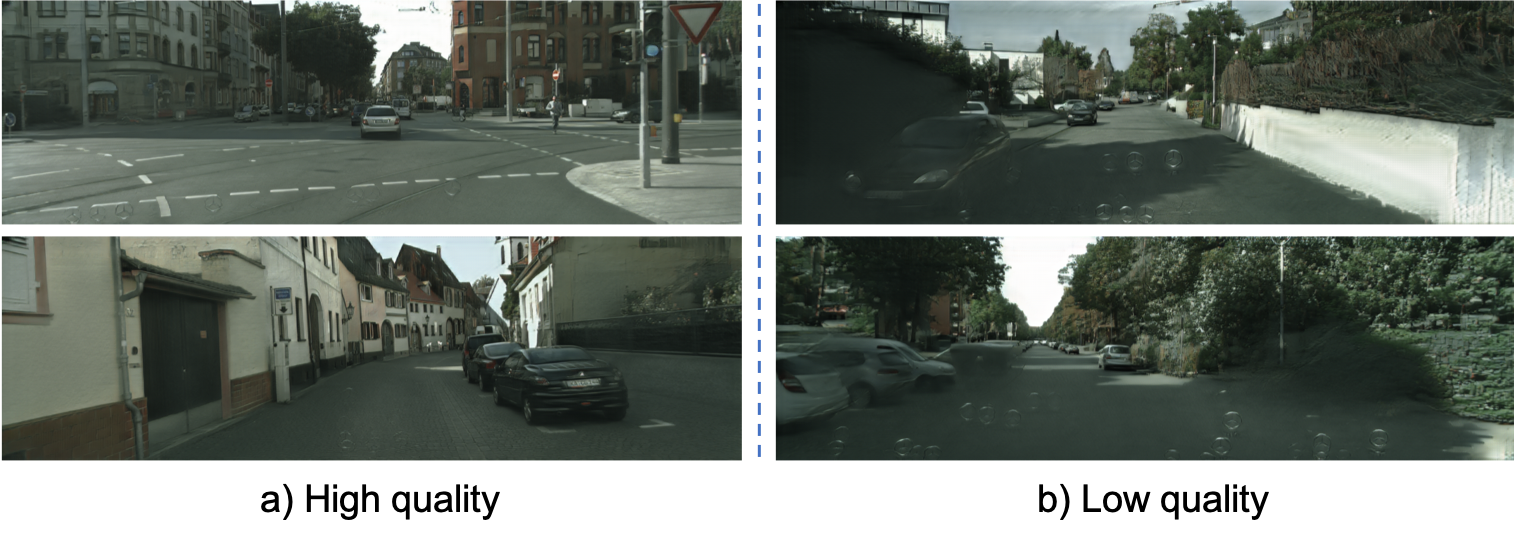}
		\end{center}
		\caption{
		Image quality examples from the KITTI dataset synthesized to be in the Cityscapes domain.
		a) shows the ones that are translated with better quality.
		Images in b) contain artifacts and fail to preserve details of the car, almost blend into the background.
  		}
		\label{fig:im_quality}
\vspace{-3mm}
\end{figure}

\vspace{-2mm}\paragraph{Weighted Supervision.}
We observe that the quality of synthetic images differs in a wide range.
For instance, some images fail to preserve details of objects or contain artifacts when translated, and these failure cases may have a larger distance to the target distribution (see Figure \ref{fig:im_quality} for an example).
This phenomenon can be also visualized in the feature space in Figure~\ref{fig:tsne_before}, where some blue dots are far away from both the source and target domains.

As a result, when performing supervised detection learning on $\mathbb{F}$ during $\mathbb{F} \rightarrow \mathbb{T}$, these defects may cause confusions to our detection model, leading to false feature alignment across domains.
To alleviate this problem, we propose an importance weighting strategy for synthetic samples based on their distances to the target distribution.
Specifically, synthetic outliers that are further away from the target distribution will receive less attention than the ones that are closer to the target domain.
We obtain the weights by taking the predicted output scores from the target domain discriminator $D_{cycle}$.
This discriminator is trained to differentiate between the source and target images with respect to the target distribution, in which the optimal discriminator is obtained with:
\begin{equation}
D^*_{cycle}(\mathbf{I}) = \frac{p_{\mathbb{T}}(\mathbf{I})}{p_{\mathbb{S}}(\mathbf{I})+p_{\mathbb{T}}(\mathbf{I})},
\end{equation}
where $\mathbf{I}$ is the synthetic target image generated via CycleGAN, and $p_{\mathbb{T}}(\mathbf{I})$ and $p_{\mathbb{S}}(\mathbf{I})$ are the probability of $\mathbf{I}$ belonging to the source and the target domain, respectively.
Here, the higher score of $D_{cycle}(\mathbf{I})$ represents a closer distribution to the target domain, thus providing a higher weight.
On the other hand, lower quality images which are further away from the target domain will be treated as outliers and receive a lower weight.
For each image $\mathbf{I}$, the importance weight is defined as:
\begin{equation}
w(\mathbf{I}) = \left\{
\begin{array}{cc}
      D_{cycle}(\mathbf{I}), & \mbox{if} \ \mathbf{I} \in \mathbb{F} \\
      1, & \mbox{otherwise.}\\
\end{array} 
\right. 
\end{equation}
We then apply this weight to the detection loss function in \eqref{eq:det_loss} when learning from synthetic images with labels during the second stage.
Thus, the final weighted objective function given images $\mathbf{I}_{\mathbb{F}}$ and $\mathbf{I}_{\mathbb{T}}$ is re-formulated based on~\eqref{eq:objective} as:
\begin{align}
  \min_{E} \max_{D} \mathcal{L}&(\mathbf{I}_\mathbb{F}, \mathbf{I}_\mathbb{T}) = w(\mathbf{I}_\mathbb{F})\mathcal{L}_{det}(\mathbf{I}_\mathbb{F}) \nonumber \\
 & + \lambda_{disc} \big[ \mathcal{L}_{disc}(E(\mathbf{I}_\mathbb{F})) + \mathcal{L}_{disc}(E(\mathbf{I}_\mathbb{T})) \big].
\end{align}

%-------------------------------------------------------------------------
\section{Experimental Results} \label{experiments}
In this section, we validate our method by evaluating the performance in three real-world scenarios that result in different domain discrepancies:
1) \textit{cross-camera adaptation},
%which can cause major visual discrepancies between domains,
%
2) \textit{weather adaptation},
and 3) \textit{adaptation to large-scale dataset}.
%adaptation from smaller annotated dataset to a much larger and diverse dataset.
%
Figure \ref{fig:det_ex} shows examples of the detection results from the three tasks before and after applying our domain adaptation method.

For each adaptation scenario, we show a baseline Faster R-CNN result trained on the source data without applying domain adaptation, and a supervised model trained fully on the target domain data (oracle) to illustrate the existing gap between domains.
Then we train the proposed model on the selected source and target domain to demonstrate the effectiveness of the proposed method.
We also conduct ablation study to analyze the effectiveness of individual proposed components.
%
%Several real-world datasets are used to conduct our study on the tasks aforementioned.
%
% In the following, we will first describe the datasets we chose and discuss the experimental results of our adaptation tasks.
%
More results will be available in the supplementary material.
All the source code and trained models will be made available to the public\footnote{\url{https://github.com/kevinhkhsu/DA_detection}}.

%-------------------------------------------------------------------------
\subsection{Implementation Details}
\paragraph{Adaptation Network.} 
In our experiments, we adopt VGG16 \cite{Simonyan14c} as the backbone for the Faster R-CNN \cite{Ren_2017} detection network, following the setting in \cite{chen2018domain}.
We design the discriminator network $D$ using 4 convolution layers with filters of size 3 $\times$ 3. 
The first 3 convolution layers have 64 channels, each followed by a leaky ReLU \cite{maas2013rectifier} with $\alpha$ set to 0.2.
The final domain classification layer has 1 channel that outputs the binary label prediction.
Our synthetic domain is generated by training CycleGAN \cite{CycleGAN2017} on the source and target domain images.

\vspace{-4mm}\paragraph{Training Details.} 
Before applying the proposed adaptation method, we pre-train the detection network using source domain images with ImageNet~\cite{deng2009imagenet} pre-trained weights.
When training the adaptation model, we use all available annotations in the source domain including the training and validation set.
%
%In our method, we first finetune the network for task $S \rightarrow F$, then further finetune on $F \rightarrow T$ to obtain final results regarding $S \rightarrow T$.
%
We optimize the network using Stochastic Gradient Descent (SGD) with a learning rate of 0.001, weight decay of 0.0005 and momentum of 0.9. We use $\lambda_{disc}=0.1$ based on a validation set to balance the discriminator loss with the detection loss. Batch size is 1 during training.
%
%We use different learning schedules among the selected scenarios and mention them separately in the corresponding experiments.
%
The proposed method is implemented with Pytorch and the networks are trained using one GTX 1080 Ti GPU with 12 GB memory.

\subsection{Datasets}
\paragraph{KITTI.} 
The KITTI dataset \cite{Geiger2012CVPR} contains images taken while driving in cities, highways, and rural areas.
There are a total of 7,481 images in the training set.
The dataset is only used as the source domain in the proposed experiments, and we utilize the full training set.

\vspace{-4mm}\paragraph{Cityscapes.} 
The Cityscapes dataset \cite{Cordts2016Cityscapes} is a collection of images with city street scenarios.
It includes instance segmentation annotation which we transform into bounding boxes for our experiments.
It contains 2,975 training images and 500 validation images.
We use Cityscapes with the KITTI dataset in Section \ref{K_C} to evaluate the cross camera adaptation and compare our results with the state-of-the-art method.

\vspace{-4mm}\paragraph{Foggy Cityscapes.} 
As self-explanatory by the name, the Foggy Cityscapes dataset \cite{SDV18} is built upon the images in the Cityscapes dataset \cite{Cordts2016Cityscapes}.
This dataset simulates the foggy weather using depth maps provided in Cityscapes with three levels of foggy weather.
The simulation process can be found in the original paper \cite{SDV18}.
Section \ref{weather} shows the experiments conducted on this simulated dataset for cross weather adaptation.

\vspace{-4mm}\paragraph{BDD100k.} 
The BDD100k dataset \cite{Yu2018BDD100KAD} consists of 100k images which are split into training, validation, and testing sets.
There are 70k training images and 10k validation images with available annotations.
This dataset includes different interesting attributes; there are 6 types of weather, 6 different scenes, 3 categories for the time of day and 10 object categories with bounding box annotation.
In our experiment, we extract a subset of the BDD100k with images labeled as $daytime$.
It includes 36,728 training and 5,258 validation images.
We use this subset to demonstrate the adaptation from a smaller dataset, Cityscapes, to a large-scale dataset using the proposed method in Section \ref{largescale}.
%These attributes allow us to conduct experiments base on the different attribute settings.
%
%We show our experiments for adapting different time of day and weather in Section \ref{timeofday} and \ref{weather} respectively.

%\subsection{KITTI $\leftrightarrow$ Cityscapes} 
\subsection{Cross Camera Adaptation} \label{K_C}
Different datasets exhibit distinct characteristics such as scenes, objects, and viewpoint.
%
%Another critical difference is the visual appearance and quality of the image.
%
%This difference is due to the underlying mechanisms and settings of the camera used to collect the dataset.
In addition, the underlying camera settings and mechanisms can also lead to critical differences in visual appearance as well as the image quality.
These discrepancies are where the domain-shift takes place.
In this experiment, we show the adaptation between images taken from different cameras and with distinctive content differences.
The KITTI \cite{Geiger2012CVPR} and Cityscapes \cite{Cordts2016Cityscapes} datasets are used as source and target respectively to conduct the cross camera adaptation experiment.
During training, all data in the KITTI training set and raw training images from Cityscapes dataset is used and further evaluated on the Cityscapes validation set.
% The adaptation network is trained for 70k iterations at stage one, then 10k more for stage two.
%
In Table \ref{tab:K_C}, we show experimental results evaluated on the $car$ class in terms of the average precision (AP).
Compared to the state-of-the-art method \cite{chen2018domain} that learns to adapt in the feature space, our baseline denoted as ``Ours (w/o synthetic)'' matches their performance using our own implementation.

\begingroup
\renewcommand{\arraystretch}{1}
\begin{table}
\begin{center}
\caption{
Cross camera adaptation using KITTI and Cityscapes datasets.
The results show the average precision (AP) of the $car$ class shared between the two domains.
}
\vspace{3mm}
\label{tab:K_C}
\begin{tabular}{lc}
				\toprule
                \multicolumn{2}{c}{KITTI $\rightarrow$ Cityscapes} \\
				\midrule
                Method  & AP \\
                \midrule
                Faster R-CNN &  28.8\\
                \midrule
                FRCNN in the wild \cite{chen2018domain} & 38.5\\
                Ours (w/o synthetic)        & 38.2 \\
                Ours (synthetic augment) & 40.6\\
                Ours (progressive) & \textbf{43.9}  \\
			   \midrule
			   Oracle & 55.8 \\
			   \bottomrule
\end{tabular}
\vspace{-3mm}
\end{center}
\end{table}
\endgroup

\begingroup
\renewcommand{\arraystretch}{0.75}
\begin{table}
\begin{center}
\caption{
Analysis of our weighted task loss compared to several arbitrary weight settings.
We show that by setting each image weight with respect to the distance from the target distribution improves the model performance.
}
\vspace{3mm}
\label{tab:weight}
\begin{tabular}{lcccccc}
				\toprule
                \multicolumn{7}{c}{KITTI $\rightarrow$ Cityscapes} \\
				\midrule
				weight & 0.8& 0.9& 1& 1.1& 1.2 & Ours\\
			   \midrule
			   AP & 39.8& 42.8& 42.2& 41.1& 42.6& \textbf{43.9}\\
			   \bottomrule
\end{tabular}
\vspace{-7mm}
\end{center}
\end{table}
\endgroup

\begingroup
\renewcommand{\arraystretch}{1}
\begin{table*}[t]
\begin{center}
\caption{
Weather adaptation focusing on clear weather to foggy weather using the Cityscapes and Foggy Cityscapes datasets respectively.
Performance is evaluated using the mean average precision (mAP) across 8 classes.
}
\vspace{1mm}
\label{tab:C_F}
\begin{tabular}{lccccccccc}
                \toprule
                                         \multicolumn{10}{c}{Cityscapes $\rightarrow$ Foggy Cityscapes} \\
                \midrule
                Method & person & rider & car & truck & bus & train & motorcycle & bicycle & mAP \\
                \midrule
                Faster R-CNN  &23.3& 29.4& 36.9& 7.1& 17.9& 2.4& 13.9& 25.7& 19.6 \\
                \midrule
                FRCNN in the wild \cite{chen2018domain} & 25.0& 31.0& 40.5& 22.1& 35.3& 20.2& 20.0& 27.1& 27.6 \\
                Diversify \& Match \cite{kim2019diversify} & 30.8&40.5 & 44.3& 27.2& 38.4& \textbf{34.5}& 28.4& 32.2& 34.6\\
                Strong-Weak Align \cite{Saito_2019_CVPR} & 29.9& 42.3& 43.5& 24.5& 36.2& 32.6& \textbf{30.0}& 35.3& 34.3\\
                Selective Align \cite{Zhu_2019_CVPR} & 33.5& 38.0& 48.5& \textbf{26.5}& 39.0& 23.3& 28.0& 33.6& 33.8\\
                Ours (w/o synthetic)    & 30.2& 37.9& 46.1& 14.7& 26.9& 7.0& 20.8& 31.5& 26.9\\
                Ours (synthetic augment)   & \textbf{36.6}& 45.3& \textbf{55.0}& 24.2& 43.9& 18.5& 28.4& \textbf{37.1}& 36.1\\
                %Ours (w/o weight) &V& & & & & & & & & \\
                Ours (progressive) & 36.0& \textbf{45.5}& 54.4& 24.3 & \textbf{44.1}& 25.8& 29.1& 35.9& \textbf{36.9}   \\        
			   \midrule
			   Oracle & 37.8& 48.4& 58.8& 25.2& 53.3& 15.8& 35.4& 39.0& 39.2\\
			   \bottomrule
\end{tabular}
\end{center}
\vspace{-7.5mm}
\end{table*}
\endgroup
In order to validate our method, we also conduct ablation studies using several settings.
First, we demonstrate the benefit of utilizing information from the synthetic domain.
When we directly augment synthetic data in the training set and include them in the source domain to perform feature-level adaptation, denoted as ``Ours (synthetic augment)'', there is a 2.1\% performance gain compared to \cite{chen2018domain}.
In the proposed method, by adopting our progressive training scheme with the importance weights,% our method takes the synthetic data as a middle ground in the feature space and connects the domain gap via progressive adaptation.
we show that our model further improves the AP by 5.4\%.
In addition, we present the advantage of our weighted task loss in balancing the uneven quality of synthetic images.
In Table \ref{tab:weight}, we show the analysis for using different fixed weights and our importance weighting method.
%
% When setting an arbitrary weight over the task loss, we obtain unstable results either with higher or lower weight.
%
Our method dynamically determines the weight of each image\footnote{In this case, the averaged weight obtained from the discriminator is around 0.9.} based on the distance from the target distribution.
Compared to the one without using any weight (i.e., weight is equal to 1), our importance weight improves the AP by 1.7\% and performs better than others that use fixed weights.
Overall, we show that our model can reduce the domain-shift problem caused by the camera along with other content differences across two distinct datasets and achieves state-of-the-art performance.

%\subsection{Cityscapes $\rightarrow$ Foggy Cityscapes}
\subsection{Weather Adaptation} \label{weather}
%cross weather adaptation
%Cityscapes, Foggy Cityscapes
%8 classes
Under real-world scenarios, supervised object detection models can be applied in different weather conditions where they may not have sufficient knowledge of.
However, it is difficult to obtain a large number of annotations in every weather condition for the models to learn.
This section studies the weather adaptation from clear weather to a foggy environment.
The Cityscapes dataset \cite{Cordts2016Cityscapes} and the Foggy Cityscapes dataset \cite{SDV18} are used as the source domain and the target domain, respectively.
%
% Since Foggy Cityscapes is directly derived from Cityscapes and thus they have similar contents, we train the model shorter for 10k iterations at the first stage and focus more at the second stage for 60k iterations.

% For a fair comparison with the state-of-the-art method \cite{chen2018domain}, we evaluate our method on 8 classes in the Cityscapes dataset as shown in Table \ref{tab:C_F}.
%
Table \ref{tab:C_F} shows that our method reduces the domain gap across weather conditions and performs favorably against the state-of-the-art methods \cite{chen2018domain,Saito_2019_CVPR,Zhu_2019_CVPR,kim2019diversify}.
%
% In addition, we discuss the characteristics of the two datasets and why it is in favor of our method.
%
When synthetic images are introduced during our progressive adaptation, there is a 10\% improvement in mAP compared to the baseline method.
We note that the target Foggy Cityscapes dataset fundamentally contain same images as the source Cityscapes dataset, but with synthesized fogs.
Thus, the synthetic target domain $\mathbb{F}$ via image translation is already closely distributed to the target domain and inherits informative labels for the network to learn.
Given such information learned from the synthetic domain, both our method and the synthetic augmented one climbs closely to the oracle result.
Although the synthetic domain lies close to the target distribution, we show in the results that our progressive training can still assist the adaptation process, improving performance and at the same time generalizing well to different categories.
To sum up, this experiment not only demonstrates the adaptation to a foggy weather condition but also shows the capability of using synthetic images to facilitate the distribution alignment process.

\begin{table*}[t]
\begin{center}
\caption{
Adaptation from a smaller Cityscapes dataset to a larger and diverse BDD100k dataset.
A subset of the BDD100k dataset labeled as $daytime$ is used as the target domain. 
We evaluate the mean average precision (mAP) of 10 classes which are available across the two domains.
}
\vspace{1mm}
\label{tab:C_B}
\begin{tabular}{lcccccccccccc}
                \toprule
                \multicolumn{12}{c}{Cityscapes $\rightarrow$ BDD100k daytime} \\
                \midrule
                Method & bike & bus & car & motor & person & rider & light & sign & train & truck & mAP \\
                \midrule
                Faster R-CNN & 19.4& 20.4& 49.0& 17.2& 31.1& 26.5& 11.5& 14.6& 0& 18.9& 20.8\\
                \midrule
                Ours (w/o synthetic) & 20.4& 20.2& 49.2& \textbf{16.6}& 32.1& 27.8& 11.9& 14.9& 0& 19.2& 21.2\\
                Ours (synthetic augment) & 23.1& \textbf{25.3}& \textbf{51.9}& 15.7& 36.0& 31.6& 12.7& 20.8& 0& \textbf{20.2}& 23.7 \\
                Ours (progressive) & \textbf{25.3}& 23.7& 51.8& 16.1& \textbf{37.6}& \textbf{32.9}& \textbf{14.0}& \textbf{22.2}& 0& 19.3& \textbf{24.3}\\        
			   \midrule
			   Oracle & 36.2& 58.2& 62.3& 36.1& 46.2& 43.6& 43.5& 49.7& 0& 57.6& 43.3\\
			  \bottomrule
			   
\end{tabular}
\end{center}
\vspace{-1mm}
\end{table*}

%BDD100k Clearday, Rainyday
%10 classes

%\subsection{Time of Day Adaptation}\label{timeofday}
%Time of day adaptation BDD100k Day, Night
%10 classes
%We choose a subset of the BDD100k images which has the attribute 'timeofday' labeled as 'day' to avoid other discrepancies. The data is then split into clear day and rainy day domains base on the 'weather' attribute.

\subsection{Adaptation to Large-scale Dataset} \label{largescale}
%cityscapes, bdd100k_daytime
%12 class
%With the fast development of digital cameras, it has never been easier to obtain a mass amount of images in the modern world.
Digital cameras have developed quickly over the years and collecting a large number of images is not a difficult task in the modern world.
%
%The difficulty in collecting image datasets appear when we want to have annotations for each of the images we obtained.
However, labeling the collected images is a major issue when it comes to building a dataset for supervised learning methods.
%
%Labels are necessary for many supervised tasks in a given domain such as object detection.
%
In this experiment, we examine the adaptation from a relatively smaller dataset to a large unlabeled domain containing distinct attributes.
We show that our method can harvest more from existing resources and adapt them to complicated environments.
To this end, we use the Cityscapes \cite{Cordts2016Cityscapes} and BDD100k \cite{Yu2018BDD100KAD} datasets as the source and target domains, respectively.
We choose a subset of the BDD100k dataset annotated as $daytime$ to be our target domain and consider the city scene as the adaptation factor, since there only exists daytime data in the Cityscapes dataset.
%
% Within the BDD100k $daytime$ image set, there are distinct attributes including numerous scenes and weather conditions.
%
% This experiment is trained with a learning rate of 0.0001 and a discriminator learning rate of 0.0005.
%
% The training schedule is 10k at the first stage and another 30k for stage two.
%

%Performance is evaluated on the 10 classes in the BDD100k test set as shown in Table \ref{tab:C_B}.
%
From the baseline and oracle results shown in Table \ref{tab:C_B}, we can observe the difficulty and the significant performance gap between the source and target domains.
Without using the synthetic data, the network has a harder time in adapting to a much diverse dataset with only 0.4\% improvement after directly aligning the source and target domains using the method in \cite{chen2018domain}.
When synthetic data is introduced to the source training set, the model learns to generalize better to the target domain and increases the performance by 2.5\%.
Finally, our method progressively adapts to the target domain by utilizing the intermediate feature space and receives an 3.1\% gain in mAP compared to the baseline method \cite{chen2018domain}.
We show in this experiment that our progressive adaptation can squeeze more juice out of the available knowledge and generalize better to a diverse environment, which is a critical issue in real-world applications.
Qualitative results are shown in Figure \ref{fig:det_ex} and more results are provided in the supplementary material.
%

%-------------------------------------------------------------------------
\begin{figure*}[t]
\centering
\begin{tabular}{c@{\hspace{0.005\linewidth}}c@{\hspace{0.005\linewidth}}c@{\hspace{0.005\linewidth}}c}

\includegraphics[width = .32\linewidth]{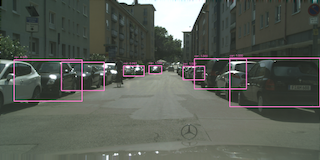} & 
\includegraphics[width = .32\linewidth]{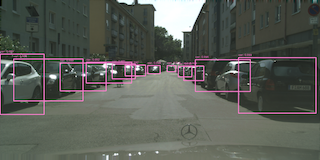} &
\includegraphics[width = .32\linewidth]{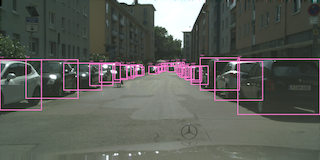} &\\

\includegraphics[width = .32\linewidth]{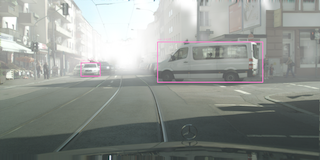} & 
\includegraphics[width = .32\linewidth]{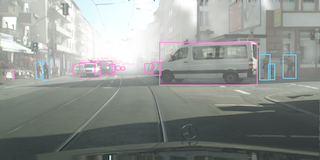} &
\includegraphics[width = .32\linewidth]{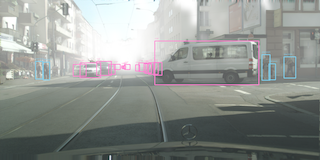} &\\

\includegraphics[width = .32\linewidth]{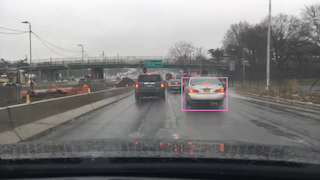} & 
\includegraphics[width = .32\linewidth]{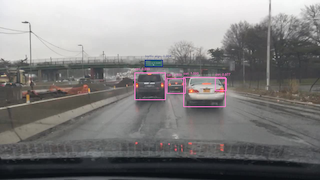} &
\includegraphics[width = .32\linewidth]{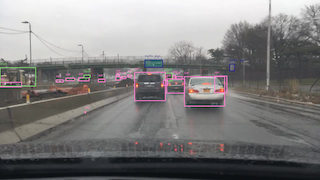} &\\

\includegraphics[width = .32\linewidth]{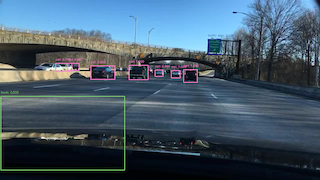} & 
\includegraphics[width = .32\linewidth]{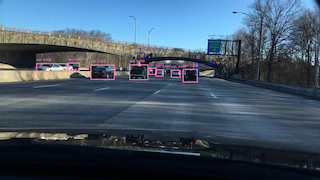} &
\includegraphics[width = .32\linewidth]{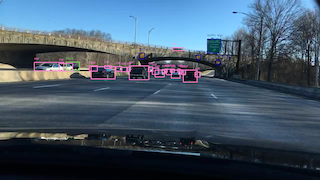} &\\
{Before Adaptation} & {After Adaptation} & {Ground Truth} & \\
\end{tabular}
\vspace{0.05em}
\caption{
Examples of the detection results from our three adaptation tasks.
The first two rows are the tasks KITTI $\rightarrow$ Cityscapes and Cityscapes $\rightarrow$ Foggy Cityscapes respectively, while the last two rows are the task Cityscapes $\rightarrow$ BDD100k.
We show the detection results on the target domain before and after applying our adaptation method as well as the ground truth labels.
}
\label{fig:det_ex}
% \vspace{15mm}
\end{figure*}
\vspace{-1mm}
\section{Conclusions}
In this paper, we propose a progressive adaptation method that bridges the domain gap using an intermediate domain, decomposing a more difficult task into two easier subtasks with a smaller gap.
We obtain the intermediate domain by transforming the source images to target ones.
Using this domain, our method progressively solves the adaptation subtasks by first adapting from source to the intermediate domain and then finally to the target domain.
In addition, we introduce a weighted loss during stage two of our method to balance different image qualities in the intermediate domain.
Experimental results show that our method performs favorably against the state-of-the-art method and can further reduce the domain discrepancy under various scenarios, such as the cross-camera case, weather condition, and adaption to a large-scale dataset.

\vspace{-2mm}
\paragraph{Acknowledgement.}
% H.-K. Hsu, C.-H. Yao, W.-C. Hung, and H.-Y. Tseng
This work is supported in part by the NSF CAREER Grant \#1149783, gifts from Adobe, Verisk, and NEC.
%-------------------------------------------------------------------------
\clearpage

{\small
\bibliographystyle{ieee}
\bibliography{egbib}
}

\end{document}